\documentclass[runningheads]{llncs}
\usepackage[T1]{fontenc}
\usepackage{graphicx,verbatim}
\usepackage{amssymb, amsmath, bm, latexsym,comment}
\usepackage{multirow}
\usepackage{xcolor}
\usepackage{array}
\usepackage{arydshln}
\usepackage[misc]{ifsym}

\providecommand{\citep}[1]{\cite{#1}}

\begin{document}
\title{CineMesh4D: Personalized 4D Whole Heart Reconstruction from Sparse Cine MRI}
\titlerunning{CineMesh4D}\flushbottom

\author{Xiaoyue Liu \inst{1} \and 
Xiaohan Yuan \inst{1,2} \and 
Mark Y Chan \inst{3,4} \and 
Ching-Hui Sia \inst{3,4} \and 
Lei Li\inst{1}${^{(\textrm{\Letter})}}$ } 
\authorrunning{Liu et al.}
\institute{Department of Biomedical Engineering, National University of Singapore, Singapore \and
School of Automation, Southeast University, Nanjing, China \and
Department of Medicine, National University of Singapore, Singapore \and
Department of Cardiology, National University Heart Centre Singapore, Singapore\\
\email{lei.li@nus.edu.sg}}

\maketitle              

\begin{abstract} 

Accurate 3D+t whole-heart mesh reconstruction from cine MRI is a clinically crucial yet technically challenging task.
The difficulty of this task arises from two coupled factors: inherently sparse sampling of 3D cardiac anatomy by 2D image slices and the tight coupling between cardiac shape and motion.
Current cardiac image-to-mesh approaches typically reconstruct only a subset of cardiac chambers or a single phase of the cardiac cycle.
In this work, we propose CineMesh4D, a novel end-to-end 4D (3D+t) pipeline that directly reconstructs patient-specific whole-heart mesh from multi-view 2D cine MRI via cross-domain mapping.
Specifically, we introduce a differentiable rendering loss that enables supervision of 3D+t whole-heart mesh from multi-view sparse contours of cine MRI.
Furthermore, we develop a dual-context temporal block that fuses global and local cardiac temporal information to capture high-dimensional sequential patterns.
In quantitative and qualitative evaluations, CineMesh4D outperforms existing approaches in terms of reconstruction quality and motion consistency, providing a practical pathway for personalized real-time cardiac assessment. 
The code will be publicly released once the manuscript is accepted.

\end{abstract} 

\section{Introduction} 

Cardiovascular diseases remain a leading cause of global mortality, underscoring the need for accurate assessment of cardiac structure and function to guide clinical decisions \cite{journal/EJIM/rivera2025}.
Cardiac digital twin, a patient-specific computational model of the heart, holds promise for personalized diagnosis and therapy \cite{journal/EHJ/thangaraj2024}. 
A critical step in building such a twin is anatomical twinning: the reconstruction of a complete 3D+t whole-heart anatomy from clinical images.
However, standard 2D cine MRI with large slice spacing provides only incomplete volumetric coverage of the heart.
This limits patient-specific anatomical context and can compromise global functional quantification.
While 3D MRI or CT offers complete coverage, their extended scan time and requirement for multiple breath hold hinder routine clinical adoption \cite{journal/CardiovascImagingAsia/kawakubo2018}. 
Reconstructing patient-specific 3D+t whole-heart anatomy directly from multi-view 2D cine MRI therefore represents a key step towards cardiac digital twins  \cite{journal/TMI/li2024}.

Recently, deep learning-based approaches have emerged as powerful paradigm for 3D cardiac geometry reconstruction \cite{journal/MIA/li2023,conference/STACOM/qiao2022}.
However, many methods are limited to ventricular geometry, reducing their applicability to whole-heart functional twinning and multi-chamber volume dynamics. 
For instance, Meng et al. \cite{journal/TMI/meng2023} proposed a prior-based deformation framework for left ventricular myocardium geometry, whereas MR-Net \cite{journal/MIA/chen2021} learned an image-mesh mapping that deformed a template mesh for biventricular anatomy.
Instead, Kong et al. \cite{journal/MedIA/kong2021} predicted whole-heart surface meshes from high-resolution 3D CT and MRI data, but they relied on 3D volumetric acquisitions which are time-consuming and resource-intensive.
In contrast, Gaggion et al. \cite{journal/MIA/gaggion2025} proposed a mesh reconstruction framework that achieved whole-heart reconstruction from multi-view sparse cine MRI.
While the framework achieves promising shape accuracy, it only considers two cardiac frames instead of the full cardiac cycle.
This prevents it from learning a joint representation of cardiac shape and motion or leveraging temporal context, which are essential for modeling cardiac dynamics.

To solve this, we develop CineMesh4D, a novel end-to-end pipeline for reconstructing high-fidelity 4D whole-heart meshes directly from sparse multi-view cine MRI.
Unlike prior approaches that rely on volumetric imaging or intermediate cardiac segmentation results, CineMesh4D learns a direct mapping from 2D+t image data to temporally coherent 3D meshes.
This is achieved through three key contributions. 
First, we propose a differentiable rendering loss derived from the Beer–Lambert law, which effectively leverages multi-view 2D+t sparse contours as supervision for 3D+t mesh reconstruction.
Second, to model reliable cardiac motion, we incorporate a dual-context temporal block that fuses global and local temporal information, producing a temporally coherent latent representation for mesh generation.
Third, the entire pipeline is trained end-to-end, jointly optimizing shape reconstruction and temporal consistency via direct image-to-mesh mapping.
This integrated framework offers a robust and practical approach for advancing patient-specific cardiac modeling and functional analysis.

\section{Methodology}

\begin{figure}[t!]
  \centering
  \includegraphics[width=\linewidth]{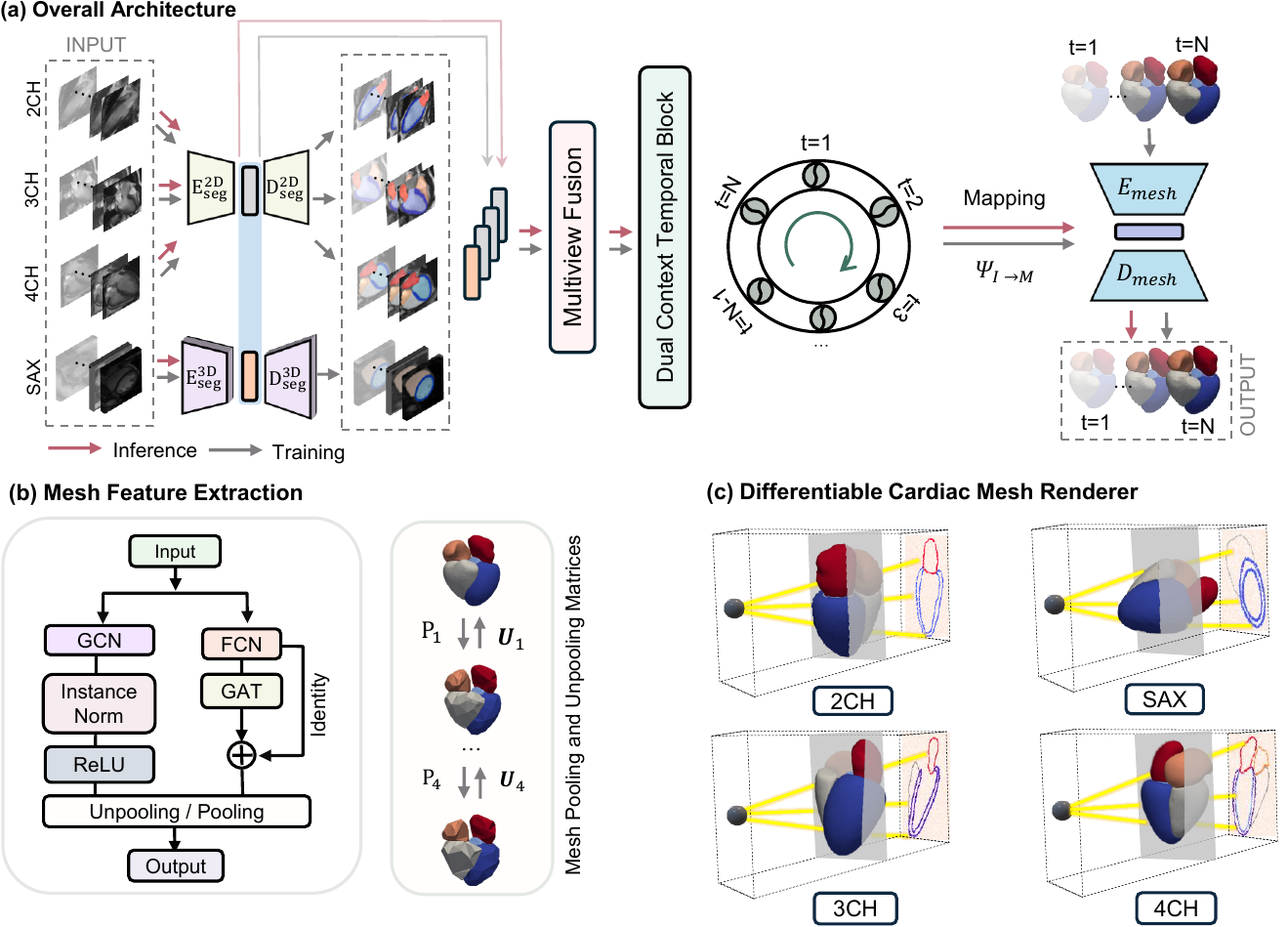}
  \caption{Framework of the proposed CineMesh4D for direct 4D whole-heart reconstruction from multi-view cine MRI via a novel image-to-mesh mapping architecture.}
  \label{fig:overall_structure}
\end{figure}

\subsection{Domain-Specific Anatomical Feature Extraction}
\label{sec:image-seg and meshVAE}
To obtain anatomical information from cine MRI, we employ a pretrained cardiac U-Net to extract the cardiac chamber regions \cite{conf/MICCAI/ronneberger2015u}.
Given multi-view cine MRI sequences $\{I_t\}_{t=1}^{N}$, we pass the CMR sequence through the pretrained U-Net encoder $E_{\text{seg}}$ to obtain anatomical feature embeddings $\mathbf{z}_t^{\text{anatomy}}\in\mathbb{R}^{d_a}$.
Specifically, we use a 2D CNN encoder $E_{\text{seg}}^{2\mathrm{D}}$ for long-axis (LAX) views and a 3D CNN encoder $E_{\text{seg}}^{3\mathrm{D}}$ for short-axis (SAX) stacks.

For the mesh domain, we adopt a variational autoencoder (VAE) to capture cardiac anatomical variability.
Each 4D mesh sequence $\{M_t\}_{t=1}^{N}$ is modeled as a sequence of spatial graphs $\{\mathcal{G}_t\}_{t=1}^{N}$, with $\mathcal{G}_t = (\mathcal{V}_t, \mathcal{E})$ representing the mesh at time $t$.
Here, $\mathcal{V}_t \in \mathbb{R}^{V \times 3}$ denotes vertex coordinates, where $V$ is the number of mesh nodes, and $\mathcal{E}$ denotes fixed mesh connectivity.
As shown in Fig. \ref{fig:overall_structure} (b), each mesh feature extraction block applies graph convolutional network (GCN) branch to embed anatomical features.
In parallel, we incorporate Exphormer \cite{conference/ICML/shirzad2023exphormer} as graph attention pathway and integrate it into our architecture via a residual skip connection.
This design complements neighborhood aggregation by enabling interactions between distant regions, strengthening structure-aware representations.
Each mesh feature extraction block concludes with a pooling matrix $\mathbf{P}_m$ or an unpooling matrix $\mathbf{U}_m$ for mesh resolution transition \cite{conference/ECCV/ranjan2018generating}.
The encoder $E_{\mathrm{mesh}}$ maps the input mesh at time step $t$ to the latent Gaussian distribution.
The mesh decoder $D_{\text{mesh}}$ reconstructs the surface as $\hat{M}_t = D_{\mathrm{mesh}}(\mathbf{z}^{\mathrm{mesh}}_t)$.
The optimization of the MeshVAE is via:
\begin{equation}
\mathcal{L}^{\mathrm{recon}}_{Mesh} =
\frac{1}{N}
\sum_{t=1}^{N}
\left\| M_t - \hat{M}_t \right\|^2
+\mathcal{L}_{\mathrm{KL}},
\end{equation}
where $M_t$ and $\hat{M}_t$ denote the input and reconstructed mesh at time $t$, and $\mathcal{L}_{\mathrm{KL}}$ is the Kullback-Leibler (KL) divergence term.

\subsection{Dual-Context Latent Regularization for Temporal Coherence} 
\begin{figure}[t!]
  \centering
  \includegraphics[width=0.9\linewidth]{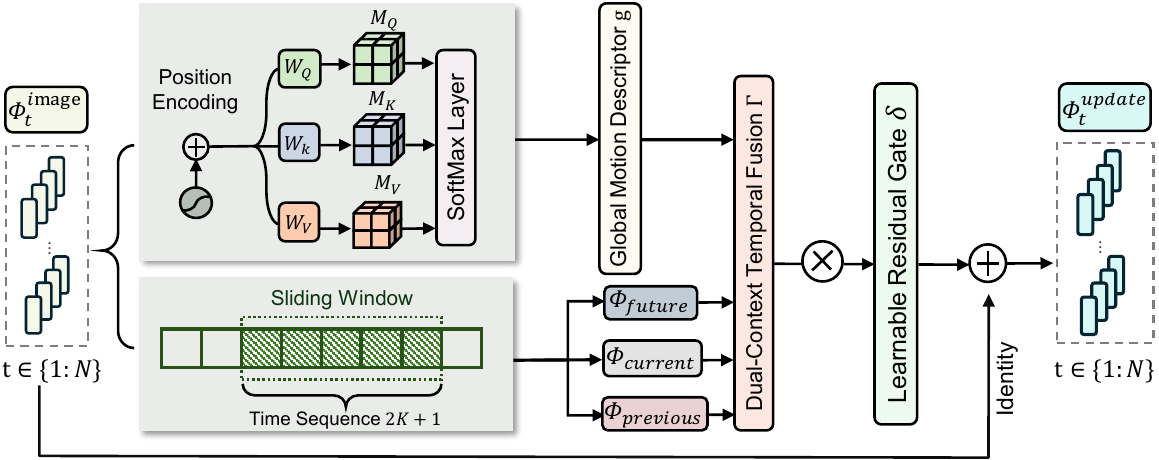}
  \caption{Dual-context temporal module to learn both global and local cardiac patterns.}
  \label{fig:ode_temporal_block}
\end{figure}

Clinically meaningful cardiac function is manifested in both cycle-level temporal patterns (global) and short-term inter-frame consistency (local). 
We therefore adopt a dual-context temporal design where the global branch summarizes the entire sequence to capture long-range temporal trends, while a local branch enforces short-term consistency among neighboring frames in the latent representation.
At each time step $t$, the anatomical latent embeddings $\mathbf{z}_t^{\text{anatomy}}$ from multiple views are passed through a multi-view fusion module implemented by fully connected layers and linear projection, yielding a unified representation $\Phi_t^{\mathrm{image}}$.
As illustrated in Fig.~\ref{fig:ode_temporal_block}, the sequential embedding is incorporated into the fused latent sequence via a sinusoidal positional encoding ~\cite{conference/neurips/vaswani2017}.
Temporal self-attention operates on the stacked latent embedding, and the pooled outputs yield a global motion descriptor $\mathbf{g}$ that encodes sequence-wide temporal context.
For each frame, neighboring frames within a temporal sliding window $\mathcal{W}_t$ of total size $(2K+1)$ are grouped into three relative temporal categories, namely $\Phi_{\mathrm{previous}}=\{\Phi_{t-j}^{\mathrm{image}}\}_{j=1}^{K}$, $\Phi_{\mathrm{current}}=\Phi_{t}^{\mathrm{image}}$, and $\Phi_{\mathrm{future}}=\{\Phi_{t+j}^{\mathrm{image}}\}_{j=1}^{K}$, for modeling direction-aware temporal dependencies.
We define $\Gamma$ as a dual-context fusion module that jointly leverages the global context $\mathbf{g}$ and the local neighborhood window $\mathcal{W}_t$ to produce the fused message $\hat{\Phi}_t$.
This is implemented by feature-space stacking of sequence-level and sliding-window context, fused by a compact projection head.
\begin{equation}
\hat{\Phi}_t = \Gamma(\mathbf{g},\mathcal{W}_t),\;
\Phi_t^{\mathrm{update}} = \Phi_t^{\mathrm{image}} + \delta \odot \hat{\Phi}_t, 
\end{equation}
Here $\delta$ is a learnable residual gating vector that controls temporal fusion.
The updated latent $\Phi_t^{\mathrm{update}}$ is then used as a temporally coherent image representation as the input of $D_{\text{mesh}}$.

\subsection{Differentiable Renderer for Contour-Guided Mesh Optimization}
\label{subsection: Differentiable Rendering}

Multi-view cine MRI captures cardiac anatomy through sparse view-specific planes, providing complementary yet inherently 2D observations of a 3D heart.
To reconstruct a complete 4D mesh under such plane supervision, a differentiable coupling is needed to translate vertex-to-plane distances into continuous per-view contributions, enabling stable contour-guided mesh optimization.
We draw inspiration from the Beer-Lambert law \cite{journal/ChemPhys/mayerhofer2020}, where light attenuates exponentially with path length: $\mathsf{I} = \mathsf{I}_0 \exp(-\mu\mathrm{L})$, and $\mu$ is the absorption coefficient. 
This offers a natural analogy: mesh vertices close to an imaging plane should contribute strongly to the projected contour, while distant vertices contribute negligibly.

As shown in Fig.~\ref{fig:overall_structure} (c), for each case and view $w \in \{\mathrm{2CH}, \mathrm{3CH}, \mathrm{4CH}, \mathrm{SAX}\}$,  we extract the affine transformation from the fixed image header to recover the view-plane geometry $\Pi^{\,w}$ in the world coordinate system.
We represent the predicted surface-mesh vertices $\hat{\mathbf{v}}_t^{\,i}$ in the same world coordinate system to ensure consistent geometric computation. Here, $i$ indexes the mesh vertices and $\hat{\mathbf{v}}_t^{\,i}$ denotes the $i$-th predicted vertex in the world coordinates.
The plane $\Pi^{\,w}$ is the view-$w$ imaging plane in world coordinates, represented by a point $\mathbf{c}^{\,w}$ on the plane and a unit normal $\mathbf{n}^{\,w}$, both in world coordinates. 
We define the vertex-to-plane normal distance as
$R_t^{\,i,w}
=\mathrm{dist}\!\left(\hat{\mathbf{v}}_t^{\,i},\,\Pi^{\,w}\right)
=\left|\mathbf{n}^{\,w}\cdot\left(\hat{\mathbf{v}}_t^{\,i}-\mathbf{c}^{\,w}\right)\right|$.
Given the vertex-to-plane distance $R_t^{\,i,w}$, we denote by $q_t^{\,i,w}$ the plane-association probability of vertex $i$ to view plane $\Pi^{\,w}$, where $\ell_{\mathrm{sigmoid}}(\cdot)$ is a sigmoid-window distance weighting and hyperparameter $\mu$ controls the sharpness of the distance-to-probability mapping, as
\(q_t^{\,i,w}=1-\exp\!\left(-\mu\,\ell_{\mathrm{sigmoid}}(R_t^{\,i,w})\right)\).
Vertex-wise plane-association probabilities $q_t^{\,i,w}$ are projected onto the corresponding view plane and aggregated to form the probability map $Q_t^{\,w}$ on each view, such that vertices closer to the imaging plane contribute more strongly, while vertices farther away are progressively down-weighted.
The differentiable rendering (DR) loss can therefore be defined as:
\begin{equation}
\mathcal{L}_{\mathrm{DR}}
=\sum_{w}\mathcal{L}_{\mathrm{B}}(Q_t^{\,w}, S_t^{\,w}),
\end{equation}
where $\mathcal{L}_{\mathrm{B}}$ follows the boundary constraint in \cite{journal/MIA/kervadec2019}, and $S_t^{\,w}$ denotes ground-truth segmentation in the view plane $w$. We compute $\mathcal{L}_{\mathrm{B}}$ at every time frame $t$.

\subsection{Optimization Objective and Inference} 

We load pretrained weights for two modality-specific pairs of encoder and decoder, $\{E_{\text{seg}}, D_{\text{seg}}\}$ for the image domain and $\{E_{\text{mesh}}, D_{\text{mesh}}\}$ for the mesh domain.
To provide additional flexibility during this process, low-rank adaptation ($\mathrm{LoRA}$) blocks are introduced into the linear layers of the mesh decoder, $\mathrm{LoRA}(D_{\text{mesh}})$, enabling parameter-efficient adaptation under a fixed backbone \cite{conf/ICLR/hu2022lora}.
Reconstruction loss $\mathcal{L}_{\mathrm{MSE}}$ is defined as the mean squared error between the predicted $\hat{M}$ and the ground-truth mesh $M$.
We apply two regularizers for mesh optimization:
$\mathcal{L}_{\mathrm{edge}}$ penalizes edge-length changes, while $\mathcal{L}_{\mathrm{norm}}$ enforces face-normal consistency \cite{conference/ECCV/wang2018}.
The optimization objective for cross-domain mapping:
\begin{equation}
\mathcal{L}_{\mathrm{Map}}=
\lambda_{\mathrm{MSE}}\,\mathcal{L}_{\mathrm{MSE}}(\hat{M},\, M)
+ \lambda_{\mathrm{DR}}\,\mathcal{L}_{\mathrm{DR}}
+ \lambda_{\mathrm{edge}}\,\mathcal{L}_{\mathrm{edge}}
+ \lambda_{\mathrm{norm}}\,\mathcal{L}_{\mathrm{norm}} 
\end{equation}
During inference, the image encoder $E_{\text{seg}}$ is frozen and image-to-mesh mapping $\Psi_{I\to M}$ conditions the mesh decoder $D_{\text{mesh}}$ to generate subject-specific meshes:
\begin{equation}
\{\hat{M}_t\}_{t=1}^{N} =
\left\{
\mathrm{LoRA}\!\left(D_{\text{mesh}}\right)
\Big(
\Psi_{I\to M}\!\big(
E_{\text{seg}}(I_t)
\big)
\Big)
\right\}_{t=1}^{N}
\end{equation}

\section{Experiments and Results}
\label{data_acquisition}

\subsubsection{Data Acquisition and Pre-Processing.} The dataset consists of 222 subjects scanned with standard multi-view cardiac cine MRI, collected from xxx, and each view comprises 25 frames.
All images were center-cropped into a unified size of $150\times150$ and fed into an automated segmentation pipeline with manual refinement \cite{conf/ICFIM/dillon2025open}. 
These refined sparse segmentations were subsequently resampled and converted into dense 3D segmentation via the atlas-deformation \cite{conf/STACOM/xu2024}. 
To generate topology-preserved whole heart mesh as reference, we further transformed the 3D segmentation onto a high-resolution template whole-heart surface mesh. 
Each mesh consists of five anatomical components, namely left ventricle (LV), LV myocardium, right ventricle (RV), left atrium (LA), and right atrium (RA).
We randomly split our dataset into 155 training, 10 validation, and 57 test cases.

\subsubsection{Implementation.} The framework was implemented in PyTorch and trained on one NVIDIA GeForce RTX 4090D GPU. 
We pretrained MeshVAE for 350 epochs using a learning rate of $1 \times 10^{-4}$ and trained the image–mesh mapping for 400 epochs using the Adam optimizer with a learning rate of $5 \times 10^{-5}$.
The temporal sliding window size is set to 5 with $K = 2$.
We set $\mu = 8$ in the rasterization loss.
Hyperparameters are selected as
$\lambda_{\mathrm{MSE}} = 10$,
$\lambda_{\mathrm{DR}} = 5$,
$\lambda_{\mathrm{edge}} = \lambda_{\mathrm{norm}}= 0.8$.

\subsubsection{Evaluation Metrics.} 
Reconstruction accuracy is quantified using vertex-wise Mean Absolute Error (MAE) and Mean Squared Error (MSE) computed between the reference and predicted surface meshes.
We report mesh-to-mesh Chamfer Distance (CD), Hausdorff Distance (HD), as well as inference time per mesh, for direct comparison with prior full heart and biventricular pipelines.
Mean Contour Distance (MCD) and Boundary F-score (BF) are reported between the reference and predicted contours for each view, where MCD measures the average contour distance and BF quantifies boundary alignment \cite{conference/CVPR/cheng2019,conference/arxiv/gur2019end}.
${E}_{{vol}}$ is reported as the whole-heart volumetric error (mL), averaged over frames, using the absolute volume difference.
Mesh jitter (${J}_{m}$) measures the temporal smoothness of vertex trajectories over time \cite{conference/CVPR/shin2024,conference/CVPR/yi2022}.

\begin{table}[t!] \centering
\caption{Quantitative evaluation of whole-heart and substructure mesh reconstruction.}
{
\fontsize{8}{11}\selectfont
\begin{tabular}{l c c c c c}
\hline
 & \textbf{Full Mesh} & \textbf{LV} & \textbf{RV} & \textbf{LA} & \textbf{RA} \\
\hline
HybridVNet & & & & & \\
\quad MAE (mm) $\downarrow$ & 2.18(0.53) & 1.70(0.54) & 1.97(0.55) & 2.31(0.70) & 2.51(0.60) \\
\quad MSE (mm$^2$) $\downarrow$ & 8.80(5.31) & 5.10(3.67) & 7.02(4.72) & 9.58(9.24) & 11.72(10.14)\\
\quad ${J}_{{m}}$ ($\mathrm{mm}/\mathrm{frame}^{3}$) $\downarrow$ & 2.29(0.23)& 1.73(0.16) & 2.07(0.15) & 2.27(0.25) & 2.70(0.28)\\
\hline
\textbf{CineMesh4D} & & & & & \\
\quad MAE (mm) $\downarrow$ & \textbf{1.68(0.31)} & \textbf{1.59(0.34)} & \textbf{1.64(0.35)} & \textbf{1.99(0.48)} & \textbf{1.86(0.58)}\\
\quad MSE (mm$^2$) $\downarrow$ & \textbf{5.06(1.79)} & \textbf{4.45(1.93)} & \textbf{4.86(2.15)} & \textbf{6.90(3.19)} & \textbf{6.26(4.06)}\\
\quad ${J}_{{m}}$ ($\mathrm{mm}/\mathrm{frame}^{3}$) $\downarrow$  & \textbf{0.77(0.17)} & \textbf{0.69(0.15)} & \textbf{0.78(0.17)} & \textbf{0.91(0.21)} & \textbf{0.93(0.23)}\\
\hline
\end{tabular} } 
\label{tab:MAE+MSE-comparison}
\end{table}

\begin{table*}[t!]
\centering
\caption{Summary of 2D contour  results of predicted mesh on different views.}

{
\fontsize{8}{11}\selectfont
\begin{tabular}{l cc cc cc}
\hline
 & \multicolumn{2}{c}{HybridVNet} & \multicolumn{2}{c}{CineMesh4D (w/o $\lambda_{\mathrm{DR}}$)} & \multicolumn{2}{c}{\textbf{CineMesh4D}} \\
\cline{2-7}
View & BF (\%) $\uparrow$ & MCD (mm) $\downarrow$ & BF (\%) $\uparrow$ & MCD (mm) $\downarrow$ & BF (\%) $\uparrow$ & MCD (mm) $\downarrow$ \\
\hline
2CH & 55.80(12.05) & 2.85(4.76) & 60.13(9.89) & 2.30(0.47) &\textbf{65.47(10.41)} & \textbf{1.99(0.41)} \\
3CH & 54.62(9.01) & 3.60(0.66) & 58.27(9.31) & 2.32(0.46) & \textbf{65.71(10.32)} & \textbf{1.97(0.44)} \\
4CH & 57.01(9.14) & 3.69(5.10) & 59.64(8.68) & 2.42(0.46) & \textbf{68.24(9.24)} & \textbf{1.89(0.38)}\\
SAX & 62.74(13.56) & 2.22(0.89) & 60.74(11.53) & 2.29(0.59) & \textbf{62.86(13.74)} & \textbf{2.03(0.62)} \\
\hline
\end{tabular}}
\label{tab:MCD_comparison}
\end{table*}

\begin{figure}[t!]
  \centering
  \includegraphics[width=0.99\linewidth]{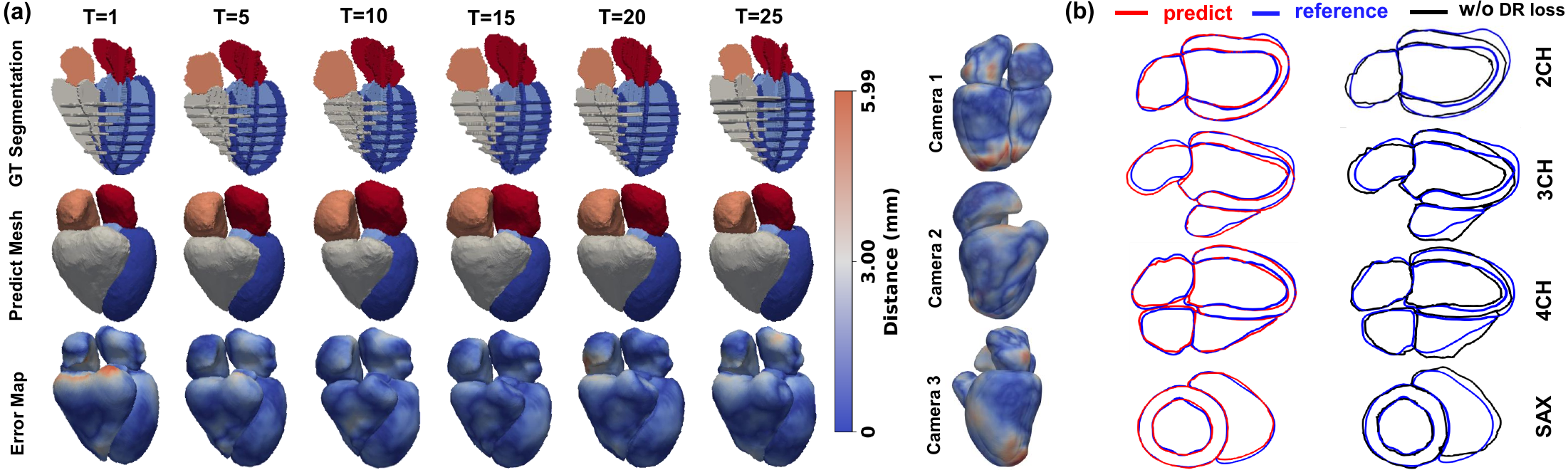}
  \caption{Visualization of reconstructed  whole heart. (a) 3D whole-heart mesh across different phases. (b) 2D contour overlap with and without differentiable
rendering (DR). }
  \label{fig:error_map}
\end{figure}

\begin{figure}[t!]
  \centering
  \includegraphics[width=\linewidth]{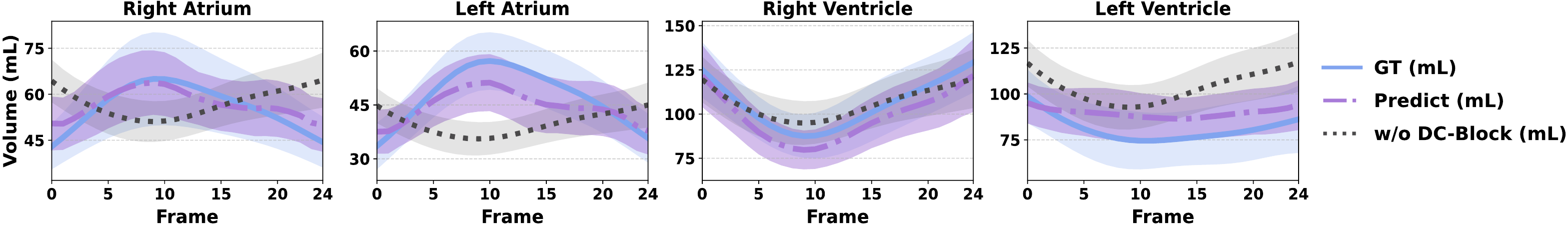}
  \caption{Illustration of cardiac chamber volume change of all test data. DC: dual-context.}
  \label{fig:fig4_per_chamber}
\end{figure}

\subsubsection{Comparison Study.} Table~\ref{tab:MAE+MSE-comparison} reports overall reconstruction errors and temporal smoothness, where CineMesh4D achieved better performance compared to the baseline across substructures. 
Ventricular reconstructions were consistently more accurate than atrial ones, as ventricles benefited from dense short-axis coverage, while atria were sparsely sampled through only a few long-axis planes. 
Table~\ref{tab:MCD_comparison} examined boundary fidelity, where BF-score increased by 17.33\%, 20.30\%, and 19.70\% relative to HybridVNet for 2CH, 3CH and 4CH views, respectively. 
Larger gains in LAX views stemmed from their single-slice nature, with only one 2D plane per view, small 3D deviations translated into pronounced contour misalignment, making direct boundary supervision through differentiable rendering particularly impactful. 
Table~\ref{tab:surface_cd_hd_time_transposed} compares surface metrics against PointNet++~\cite{conference/NIPS/qi2017}, PU-Net~\cite{conference/CVPR/yu2018}, CPD~\cite{journal/TPAMI/myronenko2010}, MR-Net~\cite{journal/MIA/chen2021}, and HybridVNet~\cite{journal/MIA/gaggion2025}, where our method achieved the lowest CD and HD. 
Qualitatively, Fig.~\ref{fig:error_map} (a) and a supplementary video visualize reconstruction quality across frames, while Fig.~\ref{fig:fig4_per_chamber} presents chamber volume curves that exhibit expected cardiac motion patterns, confirming physiological consistency of our prediction.

\begin{table}[t!]
\centering
\caption{Comparison of cardiac reconstruction results  from different methods.}
\label{tab:surface_cd_hd_time_transposed}
{\fontsize{8}{11}\selectfont
\begin{tabular}{lcccccc}
\hline
& PointNet++ & PU-Net & CPD & MR-Net & HybridVNet & \textbf{Ours} \\
\hline
CD (mm) $\downarrow$ &
13.03(2.86) & 12.15(2.85) & 12.10(6.53) & 4.39(1.38) & 4.13(1.13) & \textbf{3.41(0.62)} \\
HD (mm) $\downarrow$ &
16.04(3.57) & 15.74(3.07) & 13.05(7.04) & 6.69(1.88) & 5.17(1.02) & \textbf{5.13(0.80)} \\
Infer time (s) $\downarrow$ &
$<0.1$ & $<0.1$ & 37.45 & $<0.1$ & $<0.1$ & $<0.1$ \\
\hline
\end{tabular}
}
\end{table}

\begin{table}[t!]
\centering
\caption{Summary of the ablation study results of the proposed method.}
\label{tab:ablation_settings}
{\fontsize{8}{11}\selectfont
\setlength{\tabcolsep}{1.25pt} 
\begin{tabular}{lcccccccc}
\hline
& 
w/o SAX & 
w/o 2CH & 
w/o 3CH & 
w/o 4CH & 
w/o UNet & 
w/o VAE &
w/o DC &
\textbf{Ours}  \\
\hline
MSE$\downarrow$ & 
5.64(2.00) & 
5.59(1.94) & 
5.87(1.96) & 
5.99(1.91) & 
6.43(1.94) & 
5.67(2.03) & 
5.32(2.02)&
\textbf{5.06(1.79)} \\
MAE$\downarrow$ & 
1.76(0.31)& 
1.72(0.29) & 
1.80(0.30) & 
1.82(0.28) & 
1.87(0.28) & 
1.77(0.32) &
1.69(0.29)&
\textbf{1.68(0.31)} \\
${E}_{{vol}}$$\downarrow$& 
25.3(10.6)& 
27.3(12.8)& 
25.3(11.7)& 
26.3(9.8)& 
30.8(14.5)& 
25.0(10.8)&
24.7(13.0)&
\textbf{17.3(11.2)} \\
\hline
\end{tabular}}
\end{table}

\subsubsection{Ablation Study.}
Table~\ref{tab:ablation_settings} presents the quantitative results of our ablation study. 
Multi-view feature fusion proved essential, which is expected as SAX provided through-plane resolution while LAX views captured apical and longitudinal anatomy, together enabling complete 3D reconstruction. 
The exclusion of any single LAX view increased reconstruction error, with the 4CH view being most critical for shape accuracy, as it simultaneously visualized all four chambers.
Geometric priors from MeshVAE pretraining benefited downstream optimization by constraining reconstructions to a plausible anatomical manifold. 
Segmentation-based pretraining outperformed intensity-based self-supervision, as U-Net provided anatomical boundaries during segmentation, yielding cleaner edge-specific features critical for accurate mesh-image alignment. 
The dual-context block was critical for temporal consistency, particularly for atria where limited short-axis coverage yielded inherently sparser image features.
This can be further confirmed by the volume curves presented in Fig. \ref{fig:fig4_per_chamber}.
Finally, differentiable rendering supervision via $\mathcal{L}_{\mathrm{DR}}$ enforced contour alignment across imaging planes that vertex-wise reconstruction loss neglected, as presented in Fig. \ref{fig:error_map} (b).
While MSE minimized global vertex-wise distances, it did not penalize misalignment against specific 2D cuts. In contrast, our proposed $\mathcal{L}_{\mathrm{DR}}$ explicitly complement this by rendering and comparing against 2D segmentations.

\section{Conclusion}
We propose CineMesh4D, a temporally coherent framework for 3D+t whole-heart mesh reconstruction directly from sparse multi-view cine MRI sequences.
By introducing an end-to-end 4D image-to-mesh pipeline, a Beer–Lambert inspired differentiable rendering loss, and a dual-context temporal block, CineMesh4D achieves reliable reconstruction accuracy and strong temporal consistency.
Our work establishes a practical framework for personalized cardiac function analysis and enables downstream applications.
In future work, we will incorporate pathological metadata and clinical signals to enable comprehensive cardiac disease prediction.

\bibliographystyle{splncs04}
\bibliography{A_ref}

\end{document}